\documentclass[11pt,a4paper]{article}
\usepackage[hyperref]{emnlp-ijcnlp-2019}
\usepackage{times}
\usepackage{latexsym}
\usepackage{url}

\usepackage{listings}
\usepackage{enumitem}
\usepackage{graphicx}
\graphicspath{ {./images/} }
\usepackage{multirow}
\newcommand{\nop}[1]{}

\usepackage{scrextend}
\renewcommand{\footnotesize}{\scriptsize}
\aclfinalcopy 

\title{NeuronBlocks: Building Your NLP DNN Models Like Playing Lego}

\author{Ming Gong~$^\S$ \quad Linjun Shou~$^\S$ \quad Wutao Lin$^\S$ \quad Zhijie Sang$^\S$ \quad Quanjia Yan$^\ddag$ \\ \textbf{Ze Yang}$^\S$ \quad \textbf{Feixiang Cheng}$^\S$ \quad \textbf{Daxin Jiang}$^\S$\\
  {\small $^\S$ STCA NLP Group, Microsoft, Beijing, China} \\
  {\small $^\ddag$ Research Center for Ubiquitous Computing Systems, ICT, CAS, Beijing, China}\\
  {\small \tt \{migon, lisho, wutlin, zhsang, yaze, fecheng, djiang\}@microsoft.com}\\
  {\small \tt  yanquanjia17s@ict.ac.cn}
}

\date{}

\begin{document}
\maketitle
\begin{abstract}
Deep Neural Networks (DNN) have been widely employed in industry to address various Natural Language Processing (NLP) tasks. However, many engineers find it a big overhead when they have to choose from multiple frameworks, compare different types of models, and understand various optimization mechanisms. An NLP toolkit for DNN models with both generality and flexibility can greatly improve the productivity of engineers by saving their learning cost and guiding them to find optimal solutions to their tasks. In this paper, we introduce NeuronBlocks\footnote{Code: \url{https://github.com/Microsoft/NeuronBlocks}} \footnote{Demo:  \url{https://youtu.be/x6cOpVSZcdo}}, a toolkit encapsulating a suite of neural network modules as building blocks to construct various DNN models with complex architecture. This toolkit empowers engineers to build, train, and test various NLP models through simple configuration of JSON files. The experiments on several NLP datasets such as GLUE, WikiQA and CoNLL-2003 demonstrate the effectiveness of NeuronBlocks.
\end{abstract}

\vspace{-5pt}
\section{Introduction}
\vspace{-4pt}

Deep Neural Networks (DNN) have been widely employed in industry for solving various Natural Language Processing (NLP) tasks, such as text classification, sequence labeling, question answering, etc. However, when engineers apply DNN models to address specific NLP tasks, they often face the following challenges.

\begin{itemize}[itemsep= -0.4em,topsep = 0.3em, align=left, labelsep=-0.6em, leftmargin=1.2em]
\item Multiple DNN frameworks, including TensorFlow, PyTorch, Keras, etc. It is a big overhead to learn how to program under the frameworks.
\item Diverse and fast evolving DNN models, such as CNN, RNN, and Transformer. It takes big efforts to understand the intuition and maths behind these models.
\item Various regularization and optimization mechanisms. To tune model performance for both quality and efficiency, model developers have to gain experience in Dropout, Normalization, Mixed precision training, etc. 
\item Coding and debugging complexity. Programming under DNN frameworks requires developers to be familiar with the built-in packages and interfaces. It needs much expertise to develop, debug, and optimize code.
\item Platform compatibility. It requires extra coding work to run on different platforms, such as Linux/Windows, GPU/CPU.
\vspace{-2pt}
\end{itemize}

The above challenges often hinder the productivity of engineers, and result in less optimal solutions to their given tasks. This motivates us to develop an NLP toolkit for DNN models. Before designing this NLP toolkit, we conducted a survey among engineers and identified a spectrum of three typical personas.

\begin{itemize}
[itemsep= -0.4em,topsep = 0.3em, align=left, labelsep=-0.6em, leftmargin=1.2em]
  \item The first type of engineers prefer off-the-shelf networks. Given a specific task, they expect the toolkit to suggest several end-to-end network architectures, and then they simply focus on collecting the training data, and tuning the model parameters. They hope the whole process to be extremely agile and easy.
  \item The second type of engineers would like to build the networks by themselves. However, instead of writing each line of code from scratch, they hope the toolkit to provide a rich gallery of reusable modules as building blocks. Then they can compare various model architectures constructed by the building blocks.
  \item The last type of engineers are advanced users. They want to reuse most part of the existing networks, but for critical components, they would like to make innovations and create their own modules. They hope the toolkit to have an open infrastructure, so that customized modules can be easily plugged in.
\end{itemize}

To satisfy the requirements of all the above three personas, the NLP toolkit has to be generic enough to cover as many tasks as possible. At the same time, it also needs to be flexible enough to allow alternative network architectures as well as customized modules. Therefore, we analyzed the NLP jobs submitted to a commercial centralized GPU cluster. Table~\ref{table:gpu_cluster} showed that about 87.5\% NLP related jobs belong to a few common tasks, including sentence classification, text matching, sequence labeling, machine reading comprehension (MRC), etc. It further suggested that more than 90\% of the networks were composed of several common components, such as embedding, CNN/RNN, Transformer and so on.

\begin{table}[h!]
\centering
\small
 \begin{tabular}{c|c}
 \hline
 Tasks  & Ratio \\
 \hline
 Text matching           & 39.4\% \\
 Sentence classification & 27.3\% \\
 Sequence labeling       & 14.7\% \\
 MRC                     & 6.0\%  \\
 Others                  & 12.5\% \\
 \hline
 \end{tabular}
 \caption{\small Task analysis of NLP DNN jobs submitted to a commercial centralized GPU cluster.}
 \label{table:gpu_cluster}
 \vspace{-10pt}
\end{table}

Based on the above observations, we developed \textbf{NeuronBlocks}, a DNN toolkit for NLP tasks. The basic idea is to provide two layers of support to the engineers. The upper layer targets common NLP tasks. For each task, the toolkit contains several end-to-end network templates, which can be immediately instantiated with simple configuration. The bottom layer consists of a suite of reusable and standard components, which can be adopted as building blocks to construct networks with complex architecture. By following the interface guidelines, users can also contribute to this gallery of components with their own modules.



The technical contributions of NeuronBlocks are summarized into the following three aspects.

\begin{itemize}[itemsep= -0.4em,topsep = 0.3em, align=left, labelsep=-0.6em, leftmargin=1.2em]
\item \textbf{\textit{Block Zoo}}: categorize and abstract the most commonly used DNN components into standard and reusable blocks. The blocks within the same category can be used exchangeably.
\item \textbf{\textit{Model Zoo}}: identify the most popular NLP tasks and provide alternative end-to-end network templates (in JSON format) for each task.
\item \textbf{\textit{Platform Compatibility}}: support both Linux and Windows machines, CPU/GPU chips, as well as GPU platforms such as PAI\footnote{\url{https://github.com/Microsoft/pai}}.
\end{itemize}

\section{Related Work}
There are several general-purpose deep learning frameworks, such as TensorFlow, PyTorch and Keras, which have gained popularity in NLP community. These frameworks offer huge flexibility in DNN model design and support various NLP tasks. However, building models under these frameworks requires a large overhead of mastering these framework details. Therefore, higher level abstraction to hide the framework details is favored by many engineers.

There are also several popular deep learning toolkits in NLP, including OpenNMT~\cite{klein2017opennmt}, AllenNLP~\cite{gardner2018allennlp} etc. OpenNMT is an open-source toolkit mainly targeting 
neural machine translation or other natural language generation tasks. AllenNLP provides several pre-built models for NLP tasks, such as semantic role labeling, machine comprehension, textual entailment, etc. Although these toolkits reduce the development cost, they are limited to certain tasks, and thus not flexible enough to support new network architectures or new components.

\vspace{-2pt}

\section{Design}

\vspace{-2pt}

\begin{figure}[h]
    \centering
    \includegraphics[scale=0.5, viewport=250 50 600 480, clip=true]{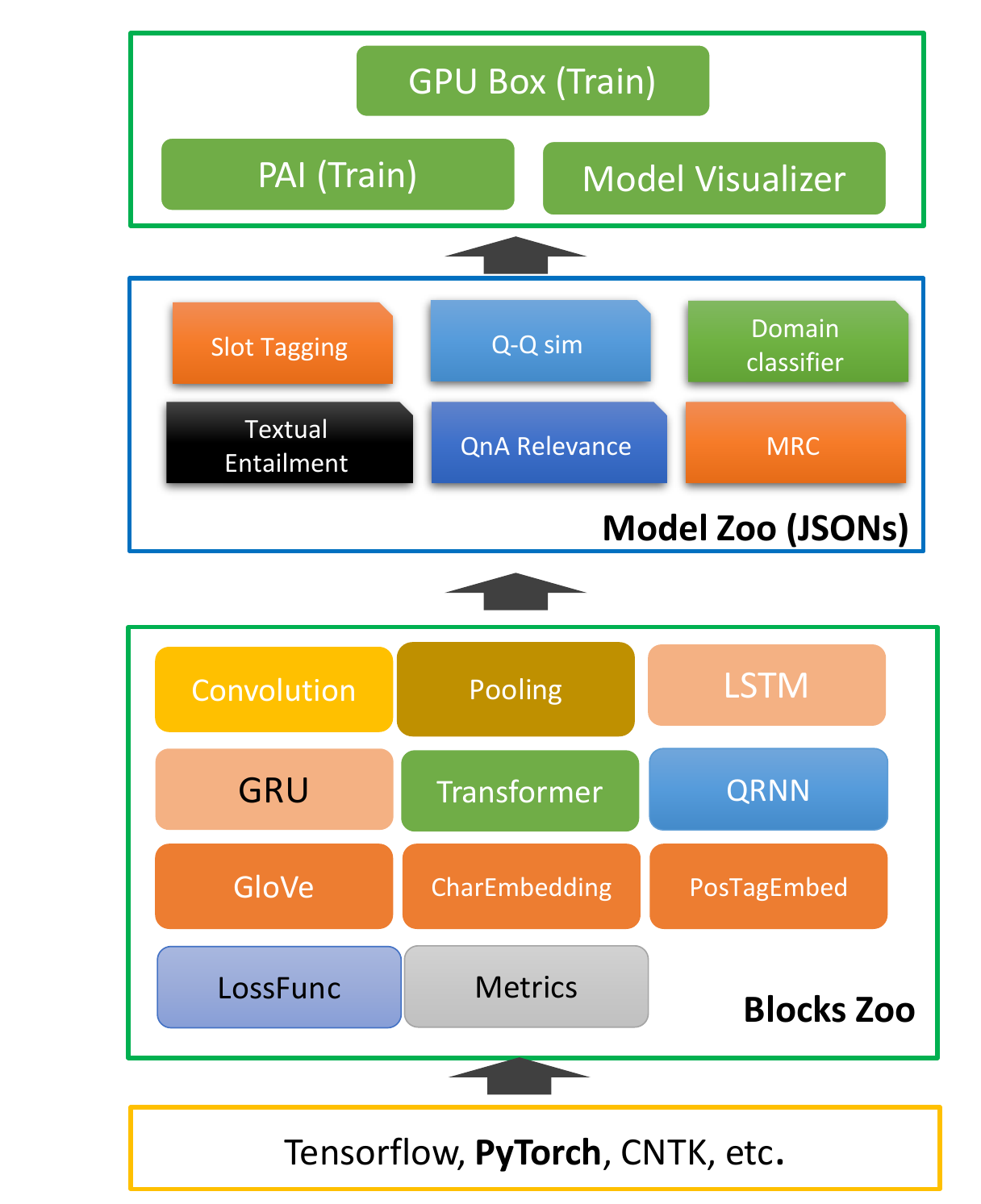}
    \caption{\small The overall framework of NeuronBlocks.}
    \vspace{-14pt}
    \label{fig:framework}
\end{figure}

Neuronblocks is built on PyTorch. The overall framework is illustrated in Figure~\ref{fig:framework}. It consists of two layers: the \textbf{Block Zoo} and the \textbf{Model Zoo}. In Block Zoo, the most commonly used components of neural networks are categorized into several groups according to their functions. Within each category, several alternative components are encapsulated into standard and reusable blocks with a consistent interface. These blocks serve as basic and exchangeable units to construct complex network architectures for different tasks. In Model Zoo, the most popular NLP tasks are identified. For each task, several end-to-end network templates are provided in the form of JSON configuration files. Users can simply browse these configurations and choose one to instantiate. The whole task can be completed without any coding efforts.


\vspace{-2pt}

\subsection{Block Zoo}

\vspace{-2pt}

We recognize the following major functional categories of neural network components. Each category covers as many commonly used modules as possible. The Block Zoo is an open framework, and more modules can be added in the future.

\begin{itemize}[itemsep= -0.4em,topsep = 0.3em, align=left, labelsep=-0.6em, leftmargin=1.2em]
\item \textbf{Embedding Layer}: Word/character embedding and extra handcrafted feature embedding such as pos-tagging are supported.

\begin{figure*}[t]
    \centering
    \includegraphics[scale=0.47, viewport=105 73 865 455, clip=true] {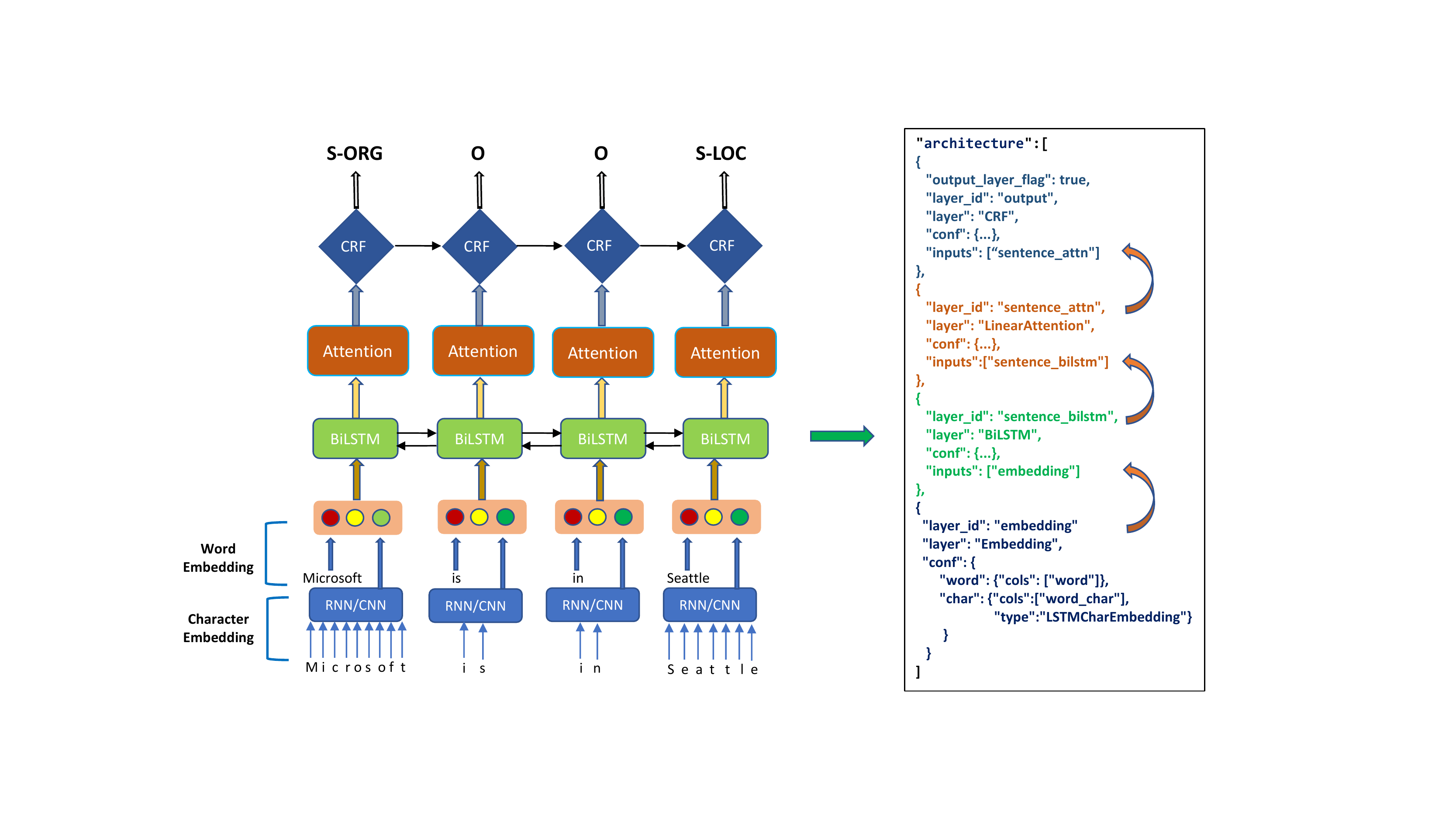}
     \vspace{-10pt}
    \caption{\small A Model architecture interface example of sequence labeling model in NeuronBlocks.}
    \label{fig:model_configuration}
\end{figure*}

\item \textbf{Neural Network Layers}: Block zoo provides common layers like RNN, CNN, QRNN~\cite{bradbury2016quasi}, Transformer~\cite{vaswani2017attention}, Highway network, Encoder Decoder architecture, etc. We also support multiple attention layers, such as Linear/Bi-linear Attention, Bidirectional attention flow~\cite{seo2016bidirectional}, etc. Meanwhile, regularization layers such as Dropout, Layer Norm, etc are also supported.

\item \textbf{Loss Function}: All built-in loss functions in PyTorch are supported. 

\item \textbf{Metrics}: For classification task, Area Under Curve (AUC), Accuracy, F1 metrics are supported. For sequence labeling task, F1/Accuracy are supported. For knowledge distillation task, MSE/RMSE are supported. For MRC task, ExactMatch/F1 are supported.
\end{itemize}

\vspace{-3pt}

\subsection{Model Zoo}

\vspace{-2pt}

In NeuronBlocks, we identify four types of most popular NLP tasks. For each task, we provide various end-to-end network templates.
\begin{itemize}[itemsep= -0.4em,topsep = 0.3em, align=left, labelsep=-0.6em, leftmargin=1.2em]
\item \textbf{Text Classification and Matching}. Tasks such as domain/intent classification, question answer matching are supported.
\item \textbf{Sequence Labeling}. Predict each token in a sequence into predefined types. Common tasks include NER, POS tagging, Slot tagging, etc.
\item \textbf{Knowledge Distillation}~\cite{DBLP:journals/corr/HintonVD15}. Teacher-Student based knowledge distillation is one common approach for model compression. NeuronBlocks provides knowledge distillation template to train light-weight student model to imitate heavy DNN models like BERT/GPT. 
\item \textbf{Extractive Machine Reading Comprehension}. Given question and passage, this task is to predict the start and end positions of the answer spans in the passage.
\end{itemize}

\begin{table*}[t!]
\centering
\small
\begin{tabular}{c|c|l|l|l|l}
\hline
\multicolumn{2}{c|}{Results(F1-score)}  & WLSTM+CRF & WLSTM & WCNN+CRF & WCNN \\
\hline
\hline
\multirow{2}{*}{Nochar} & Literature   & $89.31\pm0.1$(N) & \begin{tabular}[c]{@{}l@{}}87.00(M-16)\\  $88.49\pm0.17$(N) \end{tabular} & $88.65\pm0.2$(N)& $88.50\pm0.05$(N) \\
\cline{2-6}
& NeuronBlocks & \bf89.34 & \bf88.50 & \bf88.72 & \bf88.51 \\
\hline
\hline
\multirow{2}{*}{CLSTM}  & Literature   & \begin{tabular}[c]{@{}l@{}}90.94(L-16)\\ $91.08\pm0.08$(N)\end{tabular} & \begin{tabular}[c]{@{}l@{}}89.15(L-16)\\ $90.77\pm0.06$(N)\end{tabular} & $90.48\pm0.23$(N) & $90.28\pm0.30$(N) \\
\cline{2-6}
& NeuronBlocks & \bf91.03 & \bf90.67 & \bf90.27 & \bf90.37 \\
\hline
\hline
\multirow{2}{*}{CCNN} & Literature   & \begin{tabular}[c]{@{}l@{}}$90.91\pm0.2$(C-16)\\ 91.21(M-16)\\ $90.87\pm0.13$(P-17)\\ $91.11\pm0.21$(N)\end{tabular} & \begin{tabular}[c]{@{}l@{}}89.36(M-16)\\ $90.60\pm0.11$(N)\end{tabular} & $90.28\pm0.09$(N) & $90.51\pm0.19$(N) \\
\cline{2-6}
& NeuronBlocks & \bf91.38 & \bf90.63 & \bf90.41 & \bf90.36 \\ \hline
\end{tabular}
\vspace{-5pt}
\caption{\small NeuronBlocks results on CoNLL-2003 English NER testb dataset. The abbreviation   (C-16)=~\cite{DBLP:journals/tacl/ChiuN16},  (L-16)=~\cite{DBLP:conf/naacl/LampleBSKD16}, (M-16)=~\cite{DBLP:conf/acl/MaH16}, (N)=~\cite{DBLP:conf/coling/YangLZ18}, (P-17)=~\cite{DBLP:conf/acl/PetersABP17}. 
}
\label{table:sequence_labeling}
\end{table*}

\begin{table*}[t!]
\centering
\small
 \begin{tabular}{c|c|c|c|c|c|c|c}
 \hline
 Model & CoLA & SST-2 & QQP & MNLI & QNLI & RTE & WNLI \\
 \hline
 BiLSTM (Literature) & 17.6 & 87.5 & 85.3/82.0 & 66.7 & 77.0 & 58.5 & 56.3 \\
 +Attn (Literature) & 17.6 & 87.5 & 87.7/83.9 & 70.0 & 77.2 & 58.5 & 60.6 \\
 \hline
 BiLSTM (NeuronBlocks) & \bf 20.4 & \bf 87.5 & \bf 86.4/83.1 & \bf 69.8 & \bf 79.8 & \bf 59.2 & \bf 59.2 \\
 +Attn (NeuronBlocks) & \bf 25.1 & \bf 88.3 & \bf 87.8/83.9 & \bf 73.6 & \bf 81.0 & \bf 58.9 & \bf 59.8 \\
 \hline
 \end{tabular}
 \vspace{-5pt}
 \caption{\small NeuronBlocks�results on GLUE  benchmark development sets. As described in~\cite{wang2019glue}, for CoLA, we report Matthews correlation. For QQP, we report accuracy and F1. For MNLI, we report accuracy averaged over the matched and mismatched development sets. For all other tasks we report accuracy. All values have been scaled by 100. Please note that results on the development sets are reported, since GLUE does not distribute labels for the test sets.}
 \vspace{-15pt}
 \label{table:glue}
\end{table*}

\vspace{-4pt}

\subsection{User Interface}

NeuronBlocks provides convenient user interface\footnote{\url{https://github.com/microsoft/NeuronBlocks/blob/master/Tutorial.md}} for users to build, train, and test DNN models. The details are described in the following.

\begin{itemize}[itemsep= -0.4em,topsep = 0.3em, align=left, labelsep=-0.6em, leftmargin=1.2em]
\item  \textbf{I/O interface}. This part defines model input/output, such as training data, pre-trained models/embeddings, model saving path, etc.
\item  \textbf{Model Architecture interface}. This is the key part of the configuration file, which defines the whole model architecture. Figure \ref{fig:model_configuration} shows an example of how to specify a model architecture using the blocks in NeuronBlocks. To be more specific, it consists of a list of layers/blocks to construct the architecture, where the blocks are supplied in the gallery of \emph{Block Zoo}.
\item \textbf{Training Parameters interface}. In this part, the model optimizer as well as all other training hyper parameters are indicated. 


\end{itemize}

\begin{figure}[h]
    \centering
    \includegraphics[scale=0.4, viewport=300 130 700 450, clip=true]{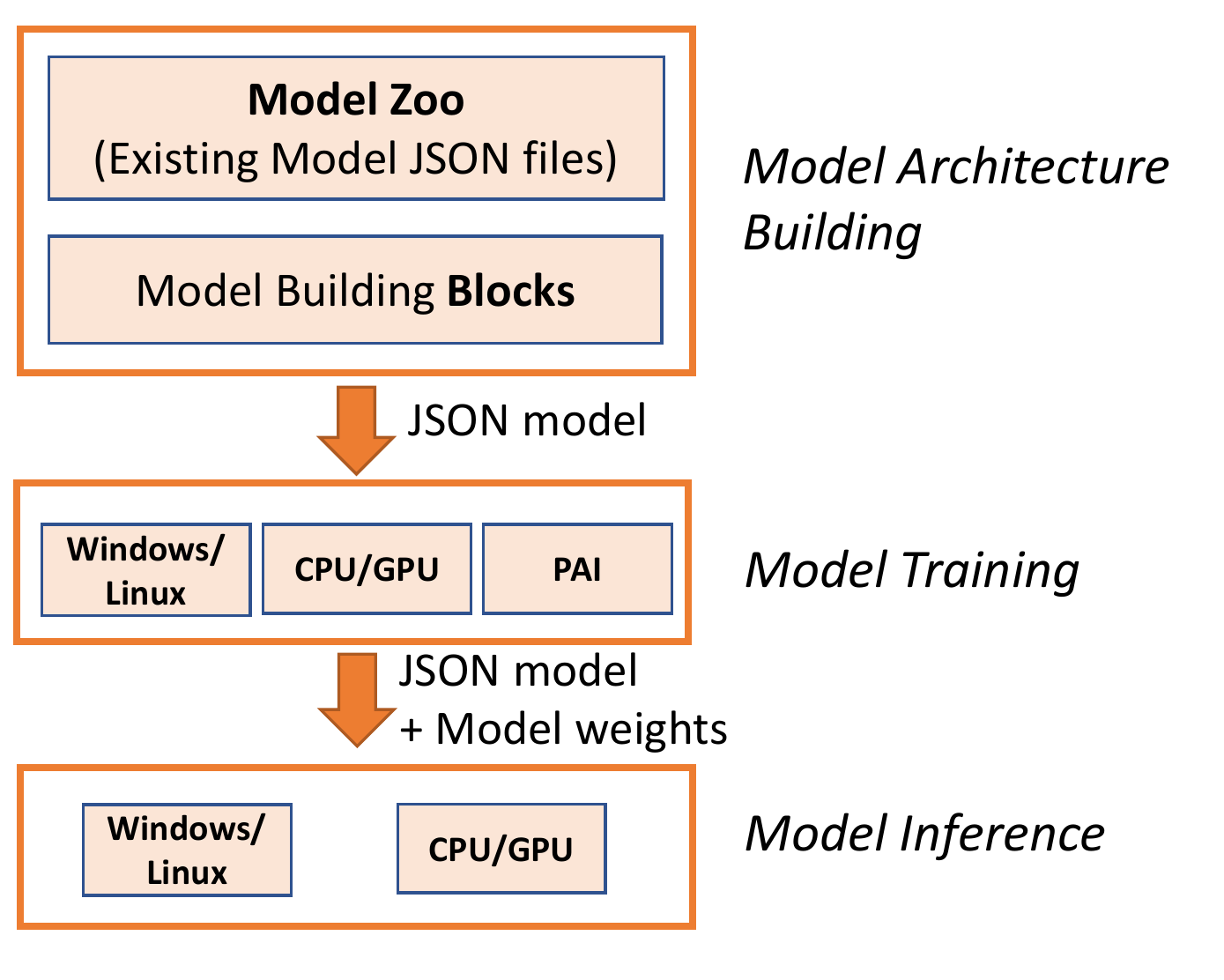}
    \caption{\small The workflow of NeuronBlocks.}
    \label{fig:workflow}
    \vspace{-20pt}
\end{figure}


\subsection{Workflow}

\begin{table*}[]
\centering
\small
\begin{tabular}{c|c|c|c}
\hline
Model & \begin{tabular}[c]{@{}c@{}}Inference Speed\\ QPS\end{tabular} & \begin{tabular}[c]{@{}c@{}}Parameters\end{tabular} & \begin{tabular}[c]{@{}c@{}}Performance\\ AUC\end{tabular} \\
\hline
Teacher Model (BERT\textsubscript{base}) & 448 & 110M  & 0.9112 \\
Student Model (BiLSTMAttn+TextCNN)  & \textbf{11128}  & \textbf{13.63M} & \textbf{0.8941} \\

\hline
\end{tabular}
\caption{\small NeuronBlocks results on Knowledge Distillation task.}
\vspace{-15pt}
\label{table:Knowledge Distillation}
\end{table*}

Figure \ref{fig:workflow} shows the workflow of building DNN models in NeuronBlocks. Users only need to write a JSON configuration file. They can either instantiate an existing template from \emph{Model Zoo}, or construct a new architecture based on the blocks from \emph{Block Zoo}. This configuration file is shared across training, test, and prediction. 

For model hyper-parameter tuning or architecture modification, users just need to change the JSON configuration file. Advanced users can also contribute novel customized blocks\footnote{\url{https://github.com/microsoft/NeuronBlocks/blob/master/Contributing.md}} into \emph{Block Zoo}, as long as they follow the same interface guidelines with the existing blocks. These new blocks can be further shared across all users for model architecture design. Moreover, NeuronBlocks has flexible platform support, such as GPU/CPU, GPU management platforms like PAI. 



\vspace{-5pt}

\section{Experiments}

\vspace{-3pt}

To verify the performance of NeuronBlocks, we conducted extensive experiments for common NLP tasks on public data sets including CoNLL-2003~\cite{DBLP:conf/conll/SangM03}, GLUE benchmark~\cite{wang2019glue},  and WikiQA corpus~\cite{yang2015wikiqa}. The experimental results showed that the models built with NeuronBlocks can achieve reliable and competitive results on various tasks, with productivity greatly improved.

\vspace{-2pt}
\subsection{Sequence Labeling}
\vspace{-3pt}
We evaluated NeuronBlocks on CoNLL-2003~\cite{DBLP:conf/conll/SangM03} English NER dataset, following most works on the same task. This dataset includes four types of named entities, namely, \emph{PERSON, LOCATION, ORGANIZATION,} and \emph{MISC}. We adopted BIOES tagging scheme instead of IOB, as many previous works indicated meaningful improvement with BIOES scheme~\cite{DBLP:conf/conll/RatinovR09}. Table~\ref{table:sequence_labeling} shows the results on CoNLL-2003 Englist testb dataset, with 12  different combinations of network layers/blocks, such as word/character embedding, CNN/LSTM and CRF. The results suggest that the flexible combination of layers/blocks in NeuronBlocks can easily reproduce the performance of original models, with comparative or slightly better performance.

\vspace{-2pt}
\subsection{GLUE Benchmark}
\vspace{-3pt}

The General Language Understanding Evaluation (GLUE) benchmark~\cite{wang2019glue} is a collection of natural language understanding tasks. We experimented on the GLUE benchmark tasks using BiLSTM and Attention based models. As shown in Table \ref{table:glue}, the models built by
NeuronBlocks can achieve competitive or even better results on GLUE tasks with minimal coding efforts. 

\vspace{-3pt}
\subsection{Knowledge Distillation}
\vspace{-3pt}

In teacher-student based knowledge distillation, a lightweight student model is trained by imitating the output of heavy teacher model like BERT. We evaluated on a binary classification dataset called Domain Classification Dataset, which was collected from a commercial search engine. Each sample in this dataset consists of two parts, i.e., a query and a binary label indicating whether the query belongs to a specific domain. Table~\ref{table:Knowledge Distillation} shows the results, where AUC is used as the evaluation criteria and Queries per Second (QPS) is used to measure inference speed. By leveraging knowledge distillation, the student model by NeuronBlocks managed to get \textbf{24.8} times inference speedup with only small performance regression compared with BERT\textsubscript{base}\footnote{\url{https://github.com/huggingface/pytorch-transformers}} fine-tuned classifier.

\vspace{-4pt}
\subsection{WikiQA}
\vspace{-2pt}
The WikiQA corpus~\cite{yang2015wikiqa} is a publicly available dataset for open-domain question answering. This dataset contains 3,047 questions from Bing query logs, each associated with some candidate answer sentences from Wikipedia. We conducted experiments on WikiQA dataset using CNN, BiLSTM, and Attention based models. The results are shown in Table~\ref{table:wikiqa}. The models built in NeuronBlocks achieved competitive or even better results with simple model configurations. 

\begin{table}[h!]
\centering
\small
 \begin{tabular}{c|c}
 \hline
 Model & AUC \\
 \hline
 CNN (~\cite{yang2015wikiqa}) & 73.59 \\
 CNN-Cnt (~\cite{yang2015wikiqa}) & 75.33 \\
 CNN (NeuronBlocks) & \bf 74.79 \\
 BiLSTM (NeuronBlocks) & \bf 76.73 \\
 BiLSTM+Attn (NeuronBlocks) & \bf 75.48 \\
 BiLSTM+MatchAttn (NeuronBlocks) & \bf 78.54 \\
 \hline
 \end{tabular}
 \caption{\small NeuronBlocks results on WikiQA.}
 \label{table:wikiqa}
 \vspace{-15pt}
\end{table}
\vspace{-3pt}

\section{Conclusion and Future Work}
\vspace{-3pt}
In this paper, we introduce NeuronBlocks, a DNN toolkit for NLP tasks built on PyTorch, targeting three types of engineers, and provides a two-layer solution to satisfy the requirements from all three types of users. To be more specific, the \emph{Model Zoo} consists of various templates for the most common NLP tasks, while the \emph{Block Zoo} supplies a gallery of alternative layers/modules for the networks. Such design achieves a balance between generality and flexibility. Extensive experiments have verified the effectiveness of this approach. NeuronBlocks has been widely used in a product team of a commercial search engine, and significantly improved the productivity for developing NLP DNN approaches. 

As an open-source toolkit, we will further extend it in various directions. The following names a few examples.

\begin{itemize}[itemsep= -0.4em,topsep = 0.3em, align=left, labelsep=-0.6em, leftmargin=1.2em]

\item Multi-task learning (MTL). In MTL, multiple related tasks are jointly trained so that knowledge learned in one task can benefit other tasks.

\item Pre-training and fine-tuning. Deep pre-training models such as ELMo~\cite{peters2018deep}, GPT~\cite{radford2018improving}, BERT~\cite{devlin2018bert} are new directions in NLP. 

\item Sequence generation task. Sequence generation is widely used in NLP fields such as machine translation~\cite{bahdanau2014neural}, text summarization~\cite{see2017get}, and dialogue systems~\cite{wen2015semantically}.

\item AutoML~\cite{elsken2019neural}. NeuronBlocks facilitates users to build models on top of \emph{Model Zoo} and \emph{Block Zoo}. With the integration of AutoML, the toolkit can further support automatic model architecture design.
\vspace{-2pt}

\end{itemize}

\vspace{-5pt}
\section{Acknowledgements}
\vspace{-3pt}

We sincerely thank the anonymous reviewers for their valuable suggestions.

\bibliography{emnlp-ijcnlp-2019}

\begin{thebibliography}{22}
\expandafter\ifx\csname natexlab\endcsname\relax\def\natexlab#1{#1}\fi

\bibitem[{Bahdanau et~al.(2015)Bahdanau, Cho, and Bengio}]{bahdanau2014neural}
Dzmitry Bahdanau, Kyunghyun Cho, and Yoshua Bengio. 2015.
\newblock Neural machine translation by jointly learning to align and
  translate.

\bibitem[{Bradbury et~al.(2017)Bradbury, Merity, Xiong, and
  Socher}]{bradbury2016quasi}
James Bradbury, Stephen Merity, Caiming Xiong, and Richard Socher. 2017.
\newblock Quasi-recurrent neural networks.

\bibitem[{Chiu and Nichols(2016)}]{DBLP:journals/tacl/ChiuN16}
Jason P.~C. Chiu and Eric Nichols. 2016.
\newblock Named entity recognition with bidirectional lstm-cnns.
\newblock \emph{{TACL}}, 4:357--370.

\bibitem[{Devlin et~al.(2019)Devlin, Chang, Lee, and
  Toutanova}]{devlin2018bert}
Jacob Devlin, Ming{-}Wei Chang, Kenton Lee, and Kristina Toutanova. 2019.
\newblock {BERT:} pre-training of deep bidirectional transformers for language
  understanding.
\newblock pages 4171--4186.

\bibitem[{Elsken et~al.(2019)Elsken, Metzen, and Hutter}]{elsken2019neural}
Thomas Elsken, Jan~Hendrik Metzen, and Frank Hutter. 2019.
\newblock Neural architecture search: A survey.
\newblock \emph{Journal of Machine Learning Research}, 20(55):1--21.

\bibitem[{Gardner et~al.(2018)Gardner, Grus, Neumann, Tafjord, Dasigi, Liu,
  Peters, Schmitz, and Zettlemoyer}]{gardner2018allennlp}
Matt Gardner, Joel Grus, Mark Neumann, Oyvind Tafjord, Pradeep Dasigi, Nelson
  Liu, Matthew Peters, Michael Schmitz, and Luke Zettlemoyer. 2018.
\newblock Allennlp: A deep semantic natural language processing platform.
\newblock \emph{arXiv preprint arXiv:1803.07640}.

\bibitem[{Hinton et~al.(2015)Hinton, Vinyals, and
  Dean}]{DBLP:journals/corr/HintonVD15}
Geoffrey~E. Hinton, Oriol Vinyals, and Jeffrey Dean. 2015.
\newblock Distilling the knowledge in a neural network.
\newblock \emph{CoRR}, abs/1503.02531.

\bibitem[{Klein et~al.(2017)Klein, Kim, Deng, Senellart, and
  Rush}]{klein2017opennmt}
Guillaume Klein, Yoon Kim, Yuntian Deng, Jean Senellart, and Alexander~M. Rush.
  2017.
\newblock Opennmt: Open-source toolkit for neural machine translation.
\newblock pages 67--72.

\bibitem[{Lample et~al.(2016)Lample, Ballesteros, Subramanian, Kawakami, and
  Dyer}]{DBLP:conf/naacl/LampleBSKD16}
Guillaume Lample, Miguel Ballesteros, Sandeep Subramanian, Kazuya Kawakami, and
  Chris Dyer. 2016.
\newblock Neural architectures for named entity recognition.
\newblock In \emph{{NAACL} {HLT} 2016, The 2016 Conference of the North
  American Chapter of the Association for Computational Linguistics: Human
  Language Technologies, San Diego California, USA, June 12-17, 2016}, pages
  260--270.

\bibitem[{Ma and Hovy(2016)}]{DBLP:conf/acl/MaH16}
Xuezhe Ma and Eduard~H. Hovy. 2016.
\newblock End-to-end sequence labeling via bi-directional lstm-cnns-crf.
\newblock In \emph{Proceedings of the 54th Annual Meeting of the Association
  for Computational Linguistics, {ACL} 2016, August 7-12, 2016, Berlin,
  Germany, Volume 1: Long Papers}.

\bibitem[{Peters et~al.(2017)Peters, Ammar, Bhagavatula, and
  Power}]{DBLP:conf/acl/PetersABP17}
Matthew~E. Peters, Waleed Ammar, Chandra Bhagavatula, and Russell Power. 2017.
\newblock Semi-supervised sequence tagging with bidirectional language models.
\newblock In \emph{Proceedings of the 55th Annual Meeting of the Association
  for Computational Linguistics, {ACL} 2017, Vancouver, Canada, July 30 -
  August 4, Volume 1: Long Papers}, pages 1756--1765.

\bibitem[{Peters et~al.(2018)Peters, Neumann, Iyyer, Gardner, Clark, Lee, and
  Zettlemoyer}]{peters2018deep}
Matthew~E. Peters, Mark Neumann, Mohit Iyyer, Matt Gardner, Christopher Clark,
  Kenton Lee, and Luke Zettlemoyer. 2018.
\newblock Deep contextualized word representations.
\newblock pages 2227--2237.

\bibitem[{Radford et~al.(2018)Radford, Narasimhan, Salimans, and
  Sutskever}]{radford2018improving}
Alec Radford, Karthik Narasimhan, Tim Salimans, and Ilya Sutskever. 2018.
\newblock Improving language understanding by generative pre-training.
\newblock \emph{URL https://s3-us-west-2. amazonaws.
  com/openai-assets/research-covers/languageunsupervised/language understanding
  paper. pdf}.

\bibitem[{Ratinov and Roth(2009)}]{DBLP:conf/conll/RatinovR09}
Lev{-}Arie Ratinov and Dan Roth. 2009.
\newblock Design challenges and misconceptions in named entity recognition.
\newblock In \emph{Proceedings of the Thirteenth Conference on Computational
  Natural Language Learning, CoNLL 2009, Boulder, Colorado, USA, June 4-5,
  2009}, pages 147--155.

\bibitem[{Sang and Meulder(2003)}]{DBLP:conf/conll/SangM03}
Erik F. Tjong~Kim Sang and Fien~De Meulder. 2003.
\newblock Introduction to the conll-2003 shared task: Language-independent
  named entity recognition.
\newblock In \emph{Proceedings of the Seventh Conference on Natural Language
  Learning, CoNLL 2003, Held in cooperation with {HLT-NAACL} 2003, Edmonton,
  Canada, May 31 - June 1, 2003}, pages 142--147.

\bibitem[{See et~al.(2017)See, Liu, and Manning}]{see2017get}
Abigail See, Peter~J. Liu, and Christopher~D. Manning. 2017.
\newblock Get to the point: Summarization with pointer-generator networks.
\newblock pages 1073--1083.

\bibitem[{Seo et~al.(2017)Seo, Kembhavi, Farhadi, and
  Hajishirzi}]{seo2016bidirectional}
Min~Joon Seo, Aniruddha Kembhavi, Ali Farhadi, and Hannaneh Hajishirzi. 2017.
\newblock Bidirectional attention flow for machine comprehension.

\bibitem[{Vaswani et~al.(2017)Vaswani, Shazeer, Parmar, Uszkoreit, Jones,
  Gomez, Kaiser, and Polosukhin}]{vaswani2017attention}
Ashish Vaswani, Noam Shazeer, Niki Parmar, Jakob Uszkoreit, Llion Jones,
  Aidan~N Gomez, {\L}ukasz Kaiser, and Illia Polosukhin. 2017.
\newblock Attention is all you need.
\newblock In \emph{Advances in neural information processing systems}, pages
  5998--6008.

\bibitem[{Wang et~al.(2019)Wang, Singh, Michael, Hill, Levy, and
  Bowman}]{wang2019glue}
Alex Wang, Amanpreet Singh, Julian Michael, Felix Hill, Omer Levy, and
  Samuel~R. Bowman. 2019.
\newblock {GLUE:} {A} multi-task benchmark and analysis platform for natural
  language understanding.
\newblock In \emph{7th International Conference on Learning Representations,
  {ICLR} 2019, New Orleans, LA, USA, May 6-9, 2019}.

\bibitem[{Wen et~al.(2015)Wen, Gasic, Mrksic, Su, Vandyke, and
  Young}]{wen2015semantically}
Tsung{-}Hsien Wen, Milica Gasic, Nikola Mrksic, Pei{-}hao Su, David Vandyke,
  and Steve~J. Young. 2015.
\newblock Semantically conditioned lstm-based natural language generation for
  spoken dialogue systems.
\newblock pages 1711--1721.

\bibitem[{Yang et~al.(2018)Yang, Liang, and Zhang}]{DBLP:conf/coling/YangLZ18}
Jie Yang, Shuailong Liang, and Yue Zhang. 2018.
\newblock Design challenges and misconceptions in neural sequence labeling.
\newblock In \emph{Proceedings of the 27th International Conference on
  Computational Linguistics, {COLING} 2018, Santa Fe, New Mexico, USA, August
  20-26, 2018}, pages 3879--3889.

\bibitem[{Yang et~al.(2015)Yang, Yih, and Meek}]{yang2015wikiqa}
Yi~Yang, Wen-tau Yih, and Christopher Meek. 2015.
\newblock Wikiqa: A challenge dataset for open-domain question answering.
\newblock In \emph{Proceedings of the 2015 Conference on Empirical Methods in
  Natural Language Processing}, pages 2013--2018.

\end{thebibliography}
\bibliographystyle{acl_natbib}

\end{document}